%% file: egpaper_for_review.tex
\documentclass[10pt,twocolumn,letterpaper]{article}

\usepackage{cvpr}
\usepackage{times}
\usepackage{epsfig}
\usepackage{graphicx}
\usepackage{amsmath}
\usepackage{amssymb}
\usepackage{enumitem}

\usepackage{comment}
\newcommand{\argmax}{\mathop{\mathrm{argmax}}}

\usepackage{color}
\newcommand{\todo}[1]{{\textcolor{red}{[#1]}}}

\newcommand{\dean}[1]{{\textcolor{blue}{[De-An: #1]}}}

\newcommand{\hide}[1]{}

\newcommand{\figref}[1]{{Figure \ref{fig:#1}}}
\newcommand{\secref}[1]{{Section \ref{sec:#1}}}
\newcommand{\eqnref}[1]{{Eq.\ (\ref{eq:#1})}}
\newcommand{\tabref}[1]{{Table \ref{tab:#1}}}

\newcommand*{\affaddr}[1]{#1} 
\newcommand*{\affmark}[1][*]{\textsuperscript{#1}}

\usepackage[pagebackref=true,breaklinks=true,letterpaper=true,colorlinks,bookmarks=false]{hyperref}

\cvprfinalcopy 

\def\cvprPaperID{790} 

\ifcvprfinal\fi

\begin{document}

\setlength{\abovedisplayskip}{7pt}
\setlength{\belowdisplayskip}{7pt}

\title{Unsupervised Visual-Linguistic Reference Resolution in Instructional Videos}

\author{De-An Huang\affmark[1], Joseph J. Lim\affmark[2], Li Fei-Fei\affmark[1], and Juan Carlos Niebles\affmark[1]\\
\affaddr{\affmark[1]Stanford University}, 
\affaddr{\affmark[2]University of Southern California}\\
{\tt\small dahuang@cs.stanford.edu, limjj@usc.edu, \{feifeili,jniebles\}@cs.stanford.edu}
}

\maketitle

\begin{abstract}
We propose an unsupervised method for reference resolution in instructional videos, where
the goal is to temporally link an entity (e.g.,\ ``dressing'') to the action (e.g.,\ ``mix yogurt'') that produced it.
The key challenge is the inevitable visual-linguistic ambiguities arising from the changes in both visual appearance and referring expression of an entity in the video. This challenge is amplified by the fact that we aim to resolve references with no supervision. We address these challenges by learning a joint visual-linguistic model, where linguistic cues can help resolve visual ambiguities and vice versa. We verify our approach by learning our model unsupervisedly using more than two thousand unstructured cooking videos from YouTube, and show that our visual-linguistic model can substantially improve upon state-of-the-art linguistic only model on reference resolution in instructional videos.
\end{abstract}

\input{intro}

\input{model}

\input{inference}
\input{result}

\input{conclusion}

\vspace{2mm}
\noindent{\bf Acknowledgement.}
This research was sponsored in part by grants from the Stanford AI Lab-Toyota Center for Artificial Intelligence Research, the Office of Naval Research (N00014-15-1-2813), and the ONR MURI (N00014-16-1-2127). 
We thank Max Wang, Rui Xu, Chuanwei Ruan, and Weixuan Gao for efforts in data collection.

{\small
\bibliographystyle{ieee}
\bibliography{egbib}
}

\end{document}

%% file: intro.tex
\section{Introduction}

\begin{figure}[tb]
\centering
   \includegraphics[width=0.94\linewidth]{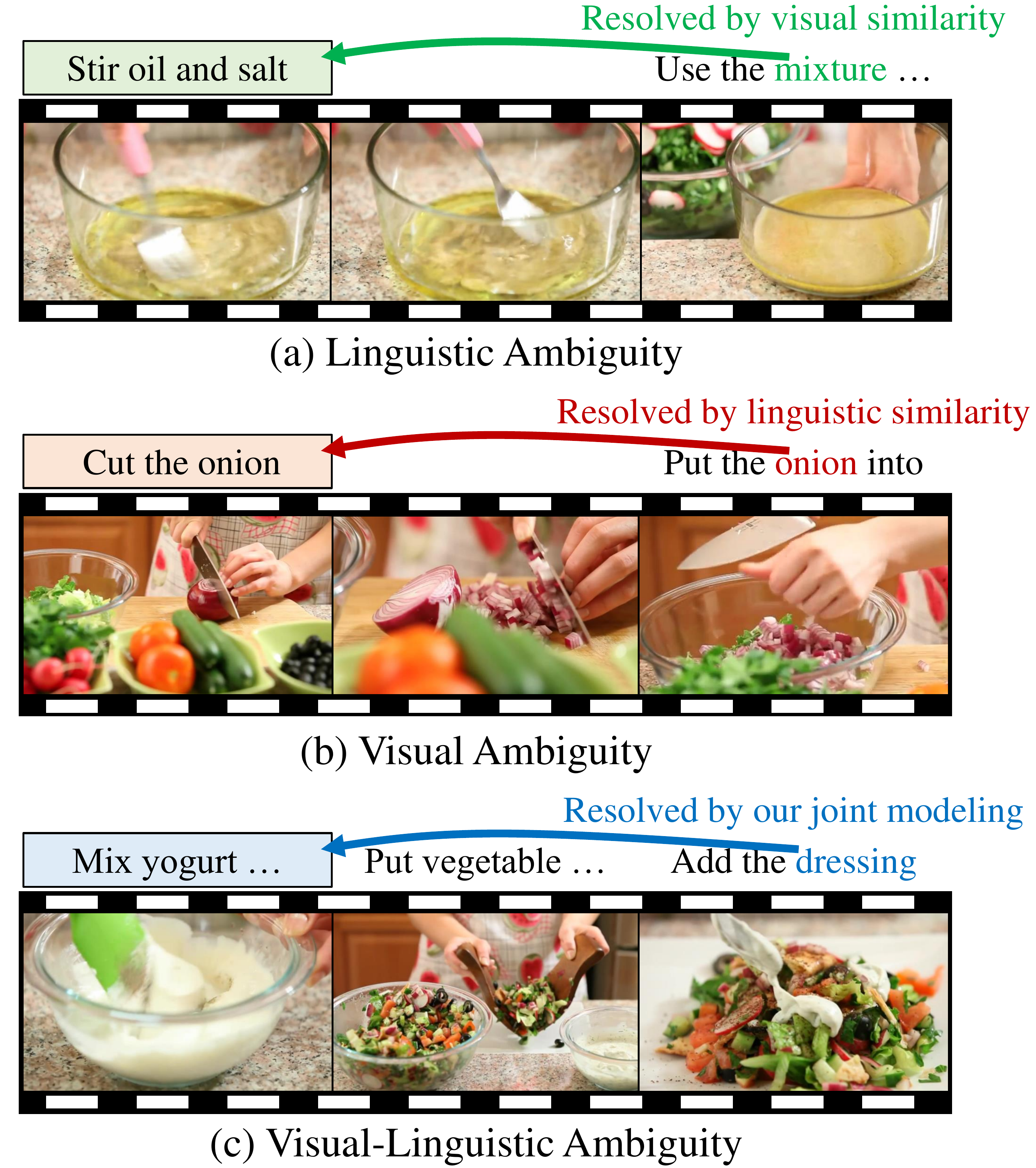}
   \caption{
   Our goal is to resolve references in videos -- temporally linking an entity to the action that produced it. (a), (b), and (c) illustrate challenges resulting from different types of ambiguities in instructional videos and how they are resolved. Our model utilizes linguistic and visual cues to resolve them. An arrow pointing to an action outcome indicates the origin of the entity.
   }
\label{fig:teaser}
\end{figure}

The number of videos uploaded to the web is growing exponentially. In this work, we are particularly interested in the narrated instructional videos.  
We as humans often acquire various types of knowledge by watching them -- from how to hold a knife to cut a tomato, to the recipe of cooking a tomato soup.
In order to build a machine with the same capability, it is necessary to understand entities (\eg knife) and actions (\eg cut) in these videos.
From a learning point of view, data from instructional videos pose a very interesting challenge. They are noisy, containing unstructured and misaligned caption uploaded by users or generated automatically by speech recognition.
Even worse, the key challenge arises from inevitable ambiguities presented in videos. For example, in \figref{teaser}(a), ``oil'' mixed with ``salt'' is later referred as a ``mixture'' -- a linguistic ambiguity due to a referring expression. An onion in \figref{teaser}(b) looks very different from its original appearance before being cut -- a visual ambiguity due to a state change. Lastly, ``yogurt'' is later referred to ``dressing'' and its appearance changes completely as shown in \figref{teaser}(c) -- both linguistic and visual ambiguities.
In this paper, we address how to resolve such ambiguities. This task is known as reference resolution: the linking of expressions to contextually given entities~\cite{schlangen2009incremental}. In other words, our goal is to extract all actions and entities from a given video, and resolve references between them. 
This is equivalent to temporally link each entity (\eg ``ice'') to the action (\eg ``freeze water'') that produced it.
For example, ``mixture'' in \figref{teaser}(a) refers to the outcome of the action ``stir oil and salt'', and ``dressing'' in \figref{teaser}(c) is the outcome of the action ``mix yogurt with black pepper''.

There have been various attempts to address reference and coreference resolution in both language understanding~\cite{bjorkelund2014learning,lee2011stanford}, and joint vision and language domains~\cite{kong2014you,liu2016jointly,plummer2015flickr30k,ramanathan2014linking}. However, most of the previous works either assume that there is enough supervision available at training time or focus on the image-sentence reference resolution, where annotations are easier to obtain. Unfortunately, obtaining high-quality reference resolution annotations in videos is prohibitively expensive and time-consuming.



Thus, in order to avoid requiring explicitly annotated data, we introduce an unsupervised method for reference resolution in instructional videos. Our model jointly learns visual and linguistic models for reference resolution -- so that it is more robust to different types of ambiguities.
Inspired by recent progress in NLP~\cite{kiddon2015mise,martschat2015latent}, we formulate our goal of reference resolution as a graph optimization task. In this case, our task of reference resolution is reformulated as finding the best set of edges (i.e. references) between nodes (i.e. actions and entities) given observation from both videos and transcriptions.


We verify our approach using unstructured instructional videos readily available on YouTube~\cite{malmaud2015s}. By jointly optimizing on over two thousand YouTube instructional videos with no reference annotation, our joint visual-linguistic model improves 9\% on both the precision and recall of reference resolution over the state-of-the-art linguistic-only model~\cite{kiddon2015mise}. We further show that resolving reference is important to aligning unstructured speech transcriptions to videos, which are usually not perfectly aligned. For a phrase like ``Cook it,'' our visual-linguistic reference model is able to infer the correct meaning of the pronoun ``it'' and improve the temporal localization of this sentence.

In summary, the main contributions of our work are: 
(1) introduce the challenging problem of reference resolution in instructional videos.
(2) propose an unsupervised graph optimization model using both visual and linguistic cues to resolve the visual and linguistic reference ambiguities.
(3) provide a benchmark for the evaluation of reference resolution in instructional videos.





\section{Related Work}

\paragraph{Coreference/Reference Resolution in Vision} In additional to the core task of coreference/reference resolution in NLP~\cite{bjorkelund2014learning,durrett2013easy,lee2011stanford}, there has been recent attempts to address these tasks in conjunction with vision. One task related to our goal of reference resolution in instructional videos is the recent progress on words to image regions reference resolution, where the goal is to spatially localize an object given a referring expression~\cite{hu2016modeling,kazemzadeh2014referitgame,krishnamurthy2013jointly,mao2015generation,nagaraja2016modeling,plummer2015flickr30k,rohrbach2015grounding,yang2016grounded,yu2016modeling}. On the other hand, coreference resolution in texts aligned with the image/video has been shown to be beneficial to the task of human naming~\cite{ramanathan2014linking}, image understanding~\cite{hodosh2010cross}, and 3D scene understanding~\cite{kong2014you}. The most related to our work is the joint optimization of name assignments to tracks and mentions in movies of Ramanathan \etal~\cite{ramanathan2014linking}. Nevertheless, our task is more challenging in both the linguistic and visual domains due to the drastic change in both visual appearances and linguistic expression introduced by state changes of the entities.



\vspace{1mm}
\noindent\textbf{Instructional Videos.} Instructional videos have been used in several contexts in computer vision. The first is semi-supervised and weakly supervised learning, where the transcription is treated as action label without accurate temporal localization~\cite{malmaud2015s,yu2014instructional}. As significant progress has been made on classifying temporally trimmed video clips, recent works aim to obtain the procedural knowledge from the instructional videos~\cite{alayrac16unsupervised,alayrac16objectstates,sener2015unsupervised}. Our goal of reference resolution in instructional videos is a step further as it requires the explicit expression of what action to act on which entities.

\vspace{1mm}
\noindent\textbf{Procedural Text Understanding.} Our goal of resolving reference in transcription of instructional videos is related to  the procedure text understanding in the NLP community~\cite{andreas2015alignment,jermsurawong2015predicting,kiddon2015mise,lau2009interpreting,long2016simpler,maeta2015framework,malmaud2014cooking}. While most approaches require supervised data (ground truth graph annotation) during training~\cite{jermsurawong2015predicting,lau2009interpreting,maeta2015framework}, Kiddon \etal proposed the first unsupervised approach for recipes interpretation~\cite{kiddon2015mise}. The linguistic part of our approach is inspired by their model. However, as we would show in the experiments, the joint modeling of language and vision plays an important role to interpret the noisier transcription in online videos.

\vspace{1mm}
\noindent\textbf{Learning from Textual Supervision.} Our learned visual model needs to observe fine-grained details in a frame based on textual supervision to improve reference resolution. This is related to recent progress on aligning and matching textual description with image~\cite{densecap,socher2014grounded} or video~\cite{bojanowski2015weakly,das2013thousand,naim2015discriminative,xu2015jointly,zhu2015aligning}. Another line of work aim to learn visual classifiers based on only textual supervision~\cite{berg2004names,bojanowski13finding,duchenne2009automatic,ramanathan2013video}. Our visual model is trained only with the transcription and is able to help reference resolution in instructional videos. 


\vspace{1mm}
\noindent\textbf{Extracting Graph from Image/Video.} 
Our formulation of reference resolution as graph optimization is related to the long-standing effort of extracting graphs from image/video. This includes recent progress in scene graphs~\cite{fidler2013sentence,johnson2015image,schuster2015generating,zitnick2013bringing}, storylines~\cite{agrawal2016sort,gupta2009understanding,huang2016visual,LinCVPR14,sigurdsson2016learning}, and action understanding~\cite{deng2016structure,pirsiavash2014parsing,soran2015generating}. Our approach of extracting  graph associating the entities with action outputs is related to works in robotics where the goal is to transform natural language instructions for the robots to execute~\cite{kollar2010toward,liu2016jointly,tellex2014learning,xiong2016robot}. It is important to note that our approach is unsupervised while a large part of the graph extraction approaches require graph annotation at the training stage.






%% file: model.tex
\section{Model}

\begin{figure}[tb]
\centering
   \includegraphics[width=0.99\linewidth]{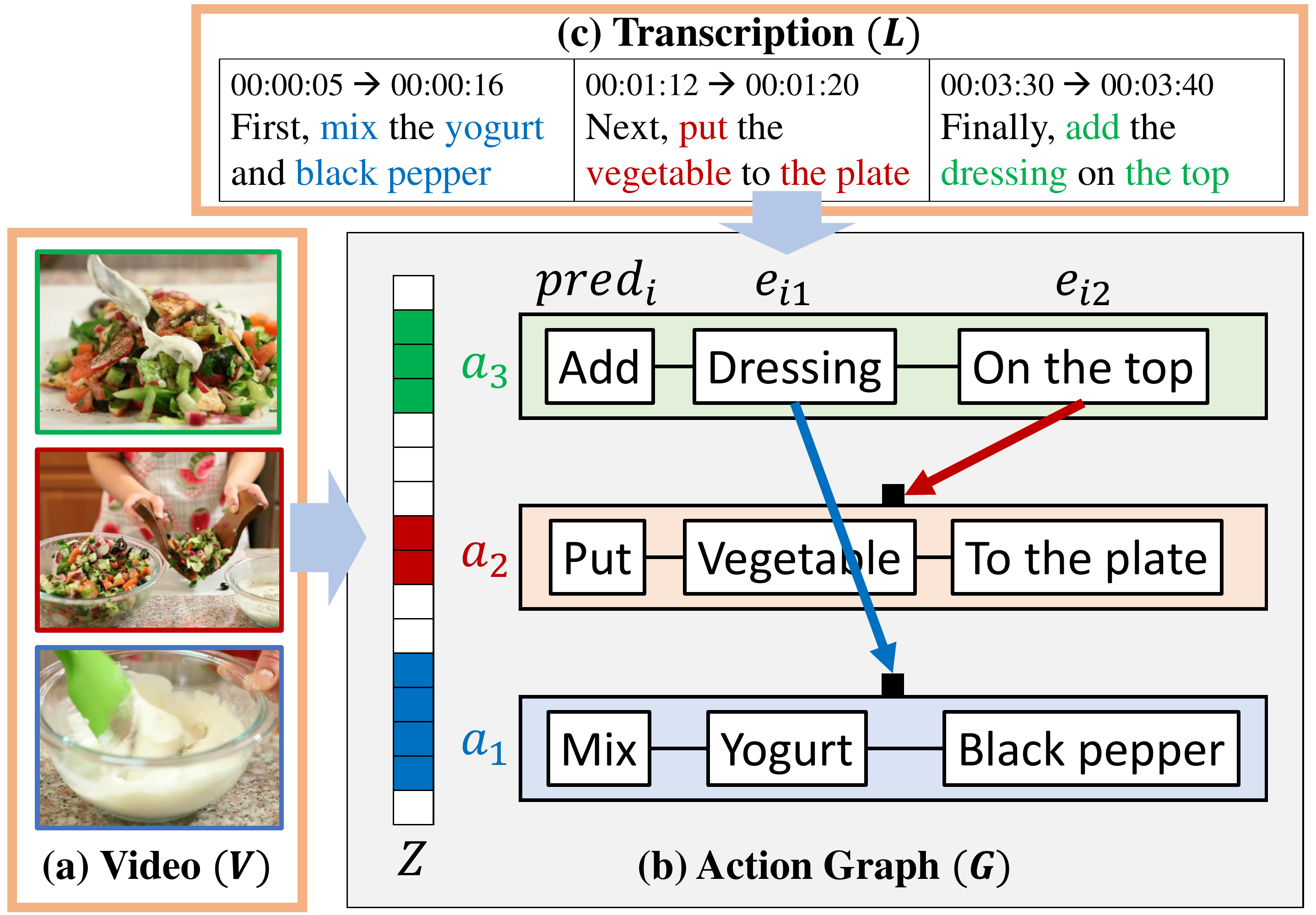}
   \caption{An action graph ($G$) is a latent representation of references in an instructional video. Both visual ($V$) and linguistic ($L$) cues of an instructional video are dependent on an action graph, and they are conditionally independent given an action graph.   
   }
\label{fig:action_graph}
\end{figure}

\hide{
\dean{Basically we'll remove $P(R|E)$ from the model, now we have very supportive results on alignment, we can talk more about it}
\dean{We get targets by parsing, and emphasize that it is very different from traditional co-reference resolution}
}

Our main goal in this paper is resolving references given an instruction video. Given a video, can we identify all references from entities to actions? For example, ``dressing'' is referring to the outcome of the action ``mix the yogurt and black pepper'' (shown in \figref{action_graph}).
Despite its many potential applications, this task comes with two major challenges. First of all, videos contain different types of ambiguities. For example, some entities change their shapes, some are referred by different names, or both. Second, obtaining a large-scale annotation for references in videos is not trivial. 

\begin{figure*}[tb]
\centering
   \includegraphics[width=0.90\linewidth]{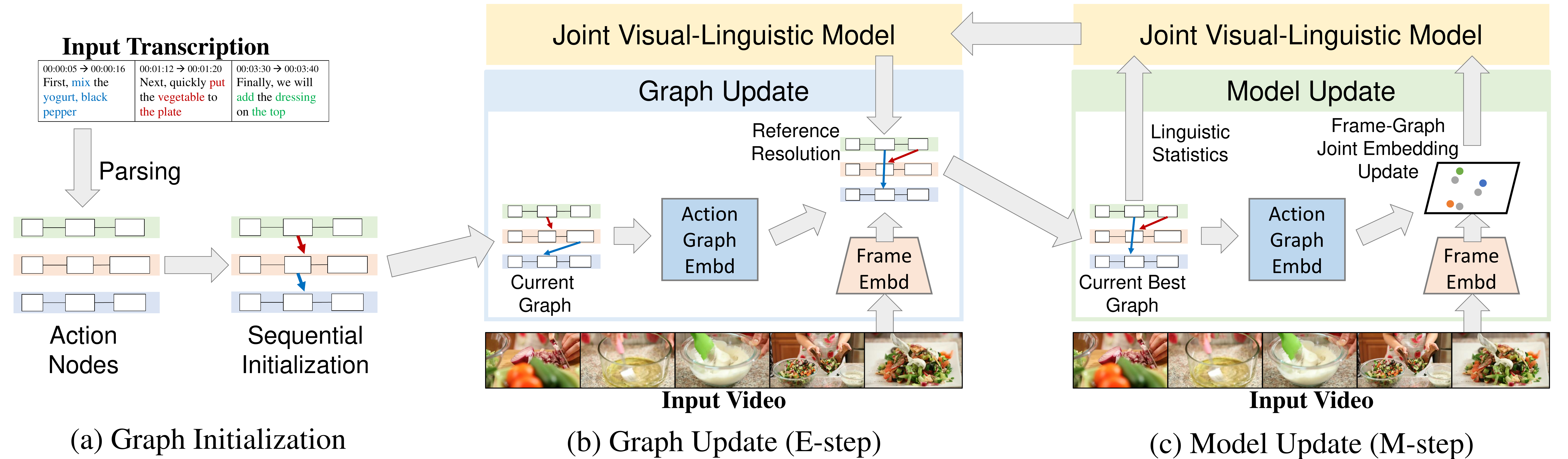}
   \caption{An overview of our optimization. (a) We first initialize the  graph by just the transcription. We alternate between (b) updating the graph with visual-linguistic reference resolution, and (c) updating the model using visual cues and linguistic statistics in the current graph}
\label{fig:system_fig}
\end{figure*}

Hence, we propose an unsupervised model for reference resolution. Our model is unique in a way that it (1) learns unsupervisedly, (2) uses both linguistic and visual cues from instructional videos, and (3) utilizes the history of actions to resolve more challenging ambiguities.
We formulate our goal of reference resolution as a graph optimization task~\cite{martschat2015latent}.
More specifically, we use the \textbf{action graph} (see \secref{action_graph})
as our latent representation because our goal of reference resolution is connecting entities to action outputs.
An overview of our unsupervised graph optimization is shown in \figref{system_fig}. We will first describe our model and discuss the details of our optimization in \secref{optimization}.






\subsection{Model Overview}

Our goal is to design an unsupervised model that can jointly learn with visual and linguistic cues of instructional videos. To this end, our model consists of a \textbf{visual model} handling video, a \textbf{linguistic model} handling transcription, and an \textbf{action graph} representation encoding all reference-related information. Our model is illustrated in \figref{action_graph}.

In summary, our task is formulated as a graph optimization task -- finding the best set of edges (\ie references) between nodes (\ie actions and entities).
Essentially, an \textit{action graph} is a latent representation of actions and their references in each video, and observations are made through a video with its visual (\ie frames) and linguistic (\ie instructions) cues; as illustrated in \figref{action_graph}. 
The fact that an \textit{action graph} contains all history information (\ie references over time) helps to resolve a complex ambiguity. Under this formulation, our approach can simply be about learning a likelihood function of an \textit{action graph} given both observations.



Formally, we optimize the following likelihood function:
\begin{equation}
\argmax_{\mathbf{G}} P(\mathbf{L}, \mathbf{V} | \mathbf{G};  \theta_V, \theta_L),
\end{equation}
where $\mathbf{G}$, $\mathbf{V}$, and $\mathbf{L}$ are the sets of temporally grounded action graph, videos, and corresponding speech transcriptions, respectively.
$\theta_V$ and $\theta_L$ are parameters of visual and linguistic models.
Under the assumption that observations are conditionally independent given the action graph, it can be further broken down into
\begin{equation}
\label{eq:whole_decomp}
\argmax_{\mathbf{G}} P(\mathbf{L} | \mathbf{G}; \theta_L)P(\mathbf{V}| \mathbf{G}; \theta_V).
\end{equation}
We can thus formulate the visual and linguistic models separately, while they are still connected via an action graph.


\subsection{Temporally Grounded Action Graph ($G$) \label{sec:action_graph}}



An \textit{action graph} is an internal representation containing all relevant information related to actions, entities, and their references: (1) action description (\eg add, dressing, on the top), (2) action time-stamp, and (3) references of entities. As an example, let's take a look at \figref{action_graph}(b), the case of making a salad. 
Each row represents an action, and each edge from an entity to an action represents a reference to the origin of the entity. Essentially, our goal is to infer these edges (\ie reference resolution).
This latent \textit{action graph} representation connects both linguistic and visual models as in \eqnref{whole_decomp}. Also, all its reference information later is used to resolve complex ambiguities, which are hard to resolve without the history of actions and references.

To this end, we define \textit{action graph} by borrowing the definition in \cite{kiddon2015mise} with a minor modification of adding temporal information. 
An action graph $G = (E,A,R)$ has $E=\{e_{ij}\}$, a set of \textbf{entity nodes} $e_{ij}$, $A=\{a_i\}$ a set of \textbf{action nodes} $a_i$ encompassing and grouping the entity nodes into actions, and $R=\{r_{ij}\}$, a set of edges corresponding to the \textbf{references} $r_{ij}$ for each entity $e_{ij}$.
The details are defined as following (See \figref{action_graph}(b) for an example):
\begin{itemize}[noitemsep,nolistsep,leftmargin=0.2in]
\vspace{3px}
	\item $a_i = (pred_i, [e_{ij}], z_i)$: action node 
    \begin{itemize}[noitemsep, nolistsep,leftmargin=0.1in]
    	\item $pred_i$: predicate or verb of the action (e.g.\ put)
        \item $e_{ij} = (t_{ij}^{syn}, t_{ij}^{sem}, S_{ij})$: entity nodes of $a_i$
        \begin{itemize}[noitemsep, nolistsep,leftmargin=0.1in]
        	\item $t_{ij}^{syn}$: its syntactic type (i.e.\ $DOBJ$ or $PP$)
            \item $t_{ij}^{sem}$: its semantic type (i.e.\ food, location, or other)
        	\item $S_{ij}$: its string representation (e.g.\ [in the bowl])
        \end{itemize}
        \item $z_i = (f^{st}, f^{end})$: starting and ending times of $a_i$
    \end{itemize}
    \item $r_{ij} = o$: directional edge or reference from entity $e_{ij}$ to its origin action node $a_o$.
\vspace{3px}
\end{itemize}

\begin{figure*}[tb]
\centering
   \includegraphics[width=0.95\linewidth]{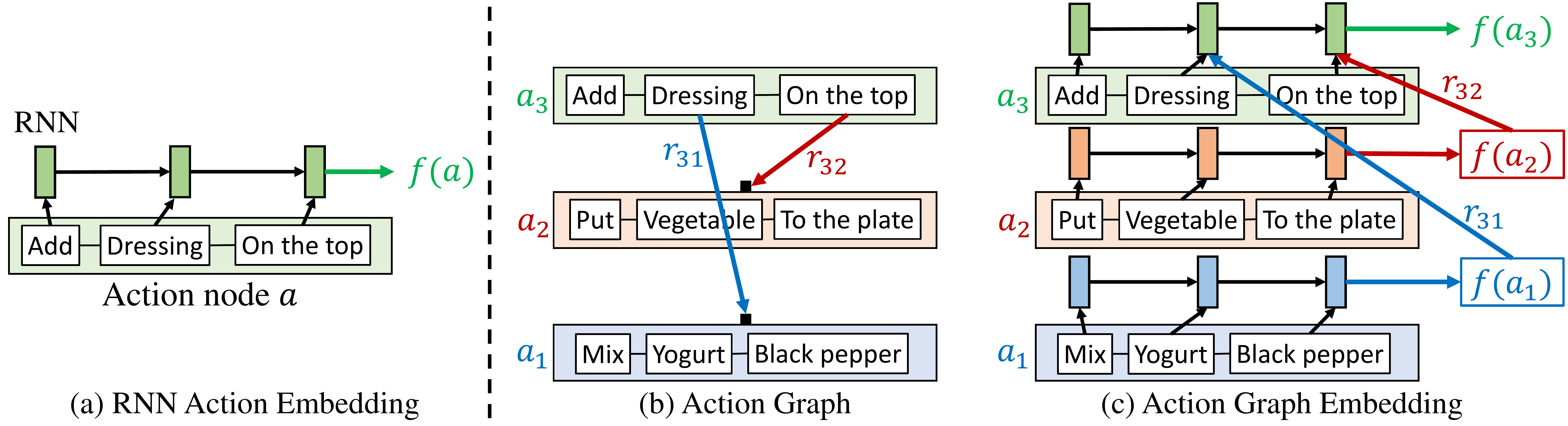}
   \caption{ 
   (a) We use RNN as the building blocks of our action graph embedding. $f(a)$ is the embedding of action $a$. (c) shows the action graph embedding of (b). In (c), the embedding of the word ``dressing'' is averaged with that of its origin, $f(a_1)$, to represent the meaning based on its reference $r_{31}$. This is then used recursively to compute $f(a_3)$, the embedding of the final step.
}
\label{fig:tree_embd}
\end{figure*}

An auxiliary action node $a_0$ is introduced for entity node not referring to the outcome of another action. For example, if the raw food entity node $e_{ij}$ ``chicken'' is not coming from another action, then $r_{ij}$ will connect $e_{ij}$ to $a_0$. In addition, we allow entity node with empty string representation $S_{ij}=[\phi]$. This can happen when the entity is \emph{implicit} in the transcription. For example, the sentence ``Add sugar'' implies an implicit entity that we can add the sugar to.


In summary, our action graph is a latent structure that constraints visual and linguistic outputs
through $P(L|G; \theta_L)$ and video $P(V|G; \theta_V)$, and also contains all reference information to resolve ambiguities. The definition of \textit{action graph} and its relationships to other models are illustrated in \figref{action_graph}. Our goal of reference resolution is reformulated as optimizing the action graph with the highest likelihood given by \eqnref{whole_decomp}.

\subsection{Visual Model}
Visual model $P(V|G;\theta_V)$ is a model that links an action graph to visual cues (\ie video frames). The motivation of our visual model is that it can help resolving linguistic-based ambiguities, and an action graph constrains visual outputs. In other words, our visual model computes a likelihood of an action graph given a set of video frames, where $\theta_V$ is the parameters of the model.



For a video $V = [x_1,\dots,x_T]$, where $x_t$ is the image frame at time $t$, and its  corresponding action graph $G$, we decompose $P(V|G; \theta_V)$ frame by frame as:
\begin{equation}
\label{eq:vm}
P(V|G; \theta_V) = \prod_{t=1}^T P(x_t|H_{\bar{z}_t})
\end{equation}
where $H_i = (a_{1:i}, r_{1:i})$ is the subgraph before action $i$, and $\bar{z}_t$ is the action label of frame $t$. That means $\bar{z}_t = i$ if frame $t$ belongs to action $i$. $\bar{z}_t = 0$ corresponds to the background.

The key novelty of our visual model is the joint formulation of frame $x_t$ and the corresponding subgraph $H_{\bar{z}_t}$. This formulation is vital to our success of improving reference resolution using visual information. Consider the final action ``add dressing on the top'' in \figref{action_graph}(b). If we swap the references of ``dressing'' and ``on the top'', then it will induce a very different meaning and thus visual appearance of this action
(i.e.\ adding vegetable on top of yogurt, instead of adding yogurt on top of vegetable). Our use of $H_{\bar{z}_t}$  instead of $a_{\bar{z}_t}$ in the visual model
catches these fine-grained differences 
and helps reference resolution; setting our approach apart from previous joint image-sentence models.
To compute $P(x_t | H_{\bar{z}_t}; \theta_V)$, we learn a joint embedding space for video frames and action (sub)graphs, inspired by visual-semantic embedding works~\cite{kiros2014unifying,socher2014grounded}.
In other words, we learn $\theta_V$ that can minimize the cosine distances between action graph features and visual frame features. 



\vspace{1mm}
\noindent\textbf{Action Graph Embedding.} In order to capture the different meanings of the action conditioned on its references, we propose a recursive definition of our action graph embedding based on RNN-based sentence embedding~\cite{kiros2015skip}. Let $g(\cdot)$ be the function of RNN embedding that takes in a list of vectors and output the final hidden state $h$. Our action graph embedding $f(\cdot)$ is recursively defined as:
\begin{equation}
f(a_i) = g\left(\left[W(pred_i), \left[W(e_{ij}) +f(a_{r_{ij}}) \right]\right]\right),
\end{equation}
where $W$ is the standard word embedding function~\cite{mikolov2013distributed,pennington2014glove}, and $r_{ij}$ is the origin of $e_{ij}$. In other words, compared to the standard sentence embedding, where the embedding of $e_{ij}$ is simply $W(e_{ij})$, we enhance it by combining with $f(a_{r_{ij}})$, the embedding of the action it is referring to. This allows our action graph embedding to capture the structure of the graph and represent different meaning of the entity based on its reference. An example is shown in \figref{tree_embd}.

\vspace{1mm}
\noindent\textbf{Frame Embedding}
We use a frame embedding function from the image captioning models~\cite{karpathy2015deep,vinyals2015show}.
By transforming the responses of convolutional layers into a vector, it has been shown to capture the fine-grained detail of the image.



%

\subsection{Linguistic Model}
Similar to the visual model, our linguistic model $P(L|G;\theta_L)$ links an action graph to linguistic observation. In our case, we use transcripts $L$ of spoken instructions in videos as our linguistic observation. Then, we know that an action graph will constrain what kind of instructions will be given in the video. Essentially, the linguistic model computes the likelihood of an action graph given transcriptions of the instructional video.

We decompose the linguistic model as follow:
\begin{align}
P(L | G; \theta_L) &= P(L, Z_L | A, R, Z ; \theta_L) \nonumber\\ 
&\propto P(L| A; \theta_L) P(A|R; \theta_L) P(Z_L| Z; \theta_L ), \label{eq:lm_final} 
\end{align}
where $Z_L$ is the time-stamps of $L$, and $A$, $R$, $Z$ are the actions, references, and time-stamps of the action graph $G$, respectively. We assume the conditional independence of the time-stamps and that $R$ is independent of $L$ given $A$.

Here, $P(L|A)$ 
parses the action nodes from transcriptions using
the Stanford CoreNLP package~\cite{manning2014stanford}. 

$P(A|R)$ measures the likelihood of the references given the actions. We adapt the model of Kiddon \etal~\cite{kiddon2015mise} and refer the readers to their paper for details. Briefly, the key models we use are:

\noindent\emph{- Verb Signature Model} to capture the property of the verb. For example, ``add'' tend to combine two food entities.

\noindent\emph{- Part-Composite Model} to represent the probable ingredients of an entity. For example, the dressing is more likely to be made up of oil compared to beef.

\noindent\emph{- Raw Food Model} to determine if an entity is an action outcome. For example, ``flour'' is less likely to be an action outcome compared to ``dough.''



We measure $P(Z_L|Z)$ independently for each action $i$, where $P(z_{Li}|z_i)$ is defined as:
\begin{equation}
\label{eq:ling_time}
P(z_{Li}|z_i)\propto e^{-\frac{|f^{st}_{Li}-f^{st}_{i}|}{\sigma}}e^{-\frac{|f^{end}_{Li}-f^{end}_{i}|}{\sigma}}
\end{equation}

%% file: inference.tex
\section{Learning \& Inference}
\label{sec:optimization}

We have discussed how we formulate references in instructional videos by the latent structure of an action graph. Using this model, our goal of reference resolution is essentially  the optimization for the most likely action graph given the videos and transcriptions based on \eqnref{whole_decomp}. 

The first challenge of optimizing \eqnref{whole_decomp} is that both the action graph $G$, and the model parameters $\theta_L$, $\theta_V$ are unknown because we aim to learn reference resolution in an unsupervised manner without any action graph annotation.

We thus take a hard EM based approach. Given the current model parameters $\theta_V$ and $\theta_L$, we estimate the temporally grounded graphs $\mathbf{G}$ (\secref{estep}). Fixing the current graphs $\mathbf{G}$, we update both the visual and linguistic models (\secref{mstep}). An overview of our optimization is shown in \figref{system_fig}.
In the following, we will describe our initialization, inference, and learning procedures in more details.

\subsection{Graph Initialization \label{sec:graph_init}}
Initially, we have neither an action graph $G$ nor model parameters $\theta_V$ and $\theta_L$. 
Hence, we initialize an action graph $G$ based on a text transcription as the following.

A list of actions $A$ is extracted using Stanford CoreNLP and the string classification model \cite{kiddon2015mise}. To simplify our task, we do not update $A$ from the initial iteration. This means all actions we consider are grounded in the transcription.
A reference $r$ of each action is initialized to one of the entities in its next action. This is proved to be a strong baseline because of the sequential nature of instructional videos~\cite{kiddon2015mise}.
A temporal location $z$ of each action is initialized as the time-stamp of the action in the transcription.

\subsection{Action Graph Optimization (E-step)}
\label{sec:estep}
In this section, we describe our approach to find the best set of action graphs $\mathbf{G}$ given model parameters $\theta_V$ and $\theta_L$. 
This is equivalent to find the best set of references $R$ and temporal groundings $Z$ for actions in each $G$, because the set of actions $A$ is fixed from initialization.
Jointly optimizing these variables is hard, and hence we relax this to finding the best $R$ and $Z$ alternatively.

Our reference optimization is based on a local search strategy \cite{kiddon2015mise}. We exhaustively update the graph with all possible swapping of two references in the current action graph, and update the graph if a reference swapped graph has a higher probability based on \eqnref{whole_decomp}. This process is repeated until there is no possible update.

To optimize our temporal alignment $Z$, we compute the probabilities of actions for each time based on a language model \eqnref{ling_time} and a visual model \eqnref{vm}. Then, we can use dynamic programming to find the optimal assignment of $Z$ to each time based on \eqnref{whole_decomp}. 

\subsection{Model Update (M-step)}
\label{sec:mstep}

Given the action graphs, we are now ready to update our linguistic and visual models.

\vspace{1mm}
\noindent\textbf{Linguistic Model Update.}
 We use the statistics of semantic and syntactic types of entities for the verb signature model. 
 For part-composite model, we use Sparse Determinant Metric Learning (SDML)\cite{qi2009efficient} to learn a metric space where the average word embedding of origin's food ingredients is close to that of the current entity $e_{ij}$. We use logistic regression to classify if the argument is a raw food. 

\vspace{1mm}
\noindent\textbf{Visual Model Update}
Given the temporally grounded action graph, for each frame $x_t$, we are able to get the corresponding subgraph $H_{\bar{z}_t}$. With it as the positive example, we collect the following negative example for our triplet loss: (1) $\tilde{H}_{\bar{z}_t}$, which is the perturbed version of $H_{\bar{z}_t}$. We randomly swap the connections in $H_{\bar{z}_t}$ to generate $\tilde{H}_{\bar{z}_t}$ as negative example. (2) $H_i$, where $i \neq \bar{z}_t$, subgraph corresponding to other frames are also negative examples. Using the positive and negative examples, we are able to update all our embeddings using backpropagation of triplet loss.

%% file: result.tex
\section{Experiments}

Given an entity such as ``dressing'', our goal is to infer its origin -- one of the previous actions. We formulate this as a graph optimization problem, where the goal is to recover the most likely references from entities to actions given the observations from transcriptions and videos. We perform the optimization \emph{unsupervisedly} with no reference supervision. In addition to our main task of reference resolution, we show that referencing is beneficial to the alignment between videos and transcriptions.

\vspace{1mm}
\noindent\textbf{Dataset.} We use the subset of $\sim$2000 videos with user uploaded caption from the WhatsCookin dataset~\cite{malmaud2015s} for our unsupervised learning.
Because there is no previous dataset with reference resolution, we annotate reference resolution labels on this subset for evaluation.
We use $k$-means clustering on the captions to select $40$ videos, and annotate action nodes $A$, their temporal locations $Z$, and references $R$. This results in $1135$ actions, more than two thousand entities and their references.
Note that this annotation is just for evaluation, and we do \emph{not} use this annotation for training.



\vspace{1mm}
\noindent\textbf{Implementation Details.}
Our visual embedding is initialized by the image captioning model of~\cite{karpathy2015deep}. Our linguistic model is initialized by the recipe interpretation model of~\cite{kiddon2015mise}. All models use learning rate $0.001$. 
For models involving both visual and linguistic parts, we always use equal weights for $P(L|G)$ and $P(V|G)$.


\subsection{Evaluating Reference Resolution}

\begin{figure*}[tb]
\centering
   \includegraphics[width=0.99\linewidth]{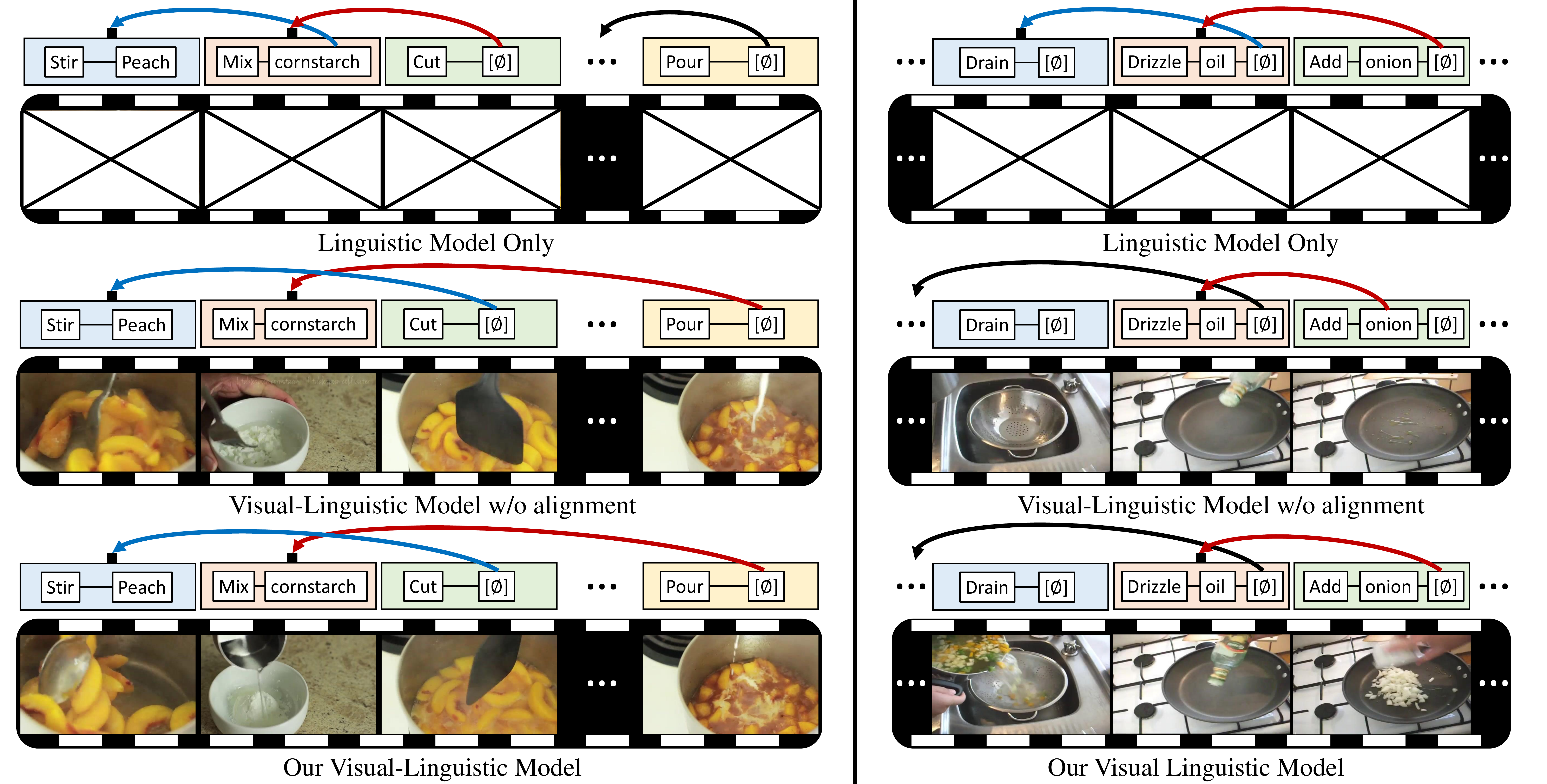}
   \caption{
Our reference resolution results. Each row shows the outputs of a type of our model. The first row is of the linguistic only model. For both videos, it fails to resolve long range references. Now, adding the visual information (the 2nd row), our model can resolve longer range references.
For example, 
in the left video, our model can correctly infer the third step is cutting peach (output two steps ahead) using the visual cue. 
Finally, we show the effect of having alignment in the process of visual-linguistic reference resolution (the 3rd row). For the right video, when the onion appears, our model recognizes that it should be another entity $\emptyset$, rather than onion, that refers ``drizzle oil''.
}
\label{fig:coref_qual}
\end{figure*}

\noindent\textbf{Experimental Setup.} 
For evaluation, we first run our model unsupervisedly on all the instructional videos in the dataset. 
The action and entity nodes here are generated automatically by the Stanford CoreNLP parser~\cite{manning2014stanford}. The semantic types of the entities are obtained using unsupervised string classification~\cite{kiddon2015mise}. After the optimization is finished, we apply one E-step of the final model to the evaluation set. In this case, we use the action and entity nodes provided by the annotations to isolate the errors introduced by the automatic parser and focus on evaluating the reference resolution in the evaluation set. We use the standard precision, recall, and F1 score as evaluation metric~\cite{kiddon2015mise}.

\vspace{1mm}
\noindent\textbf{Baselines.} We compare to the following models:

{\noindent \it  - Sequential Initialization.} This baseline seeks for the nearest preceding action that is compatible for reference resolution, which is a standard heuristic in coreference resolution. This is used as the initial graph for all the other methods.

{\noindent \it  - Visual/Linguistic Model Only.} We evaluate in separation the contribution of our visual and linguistic model. Our linguistic model is adapted from~\cite{kiddon2015mise}. We additionally incorporate word embedding and metric learning to improve its performance in instructional videos.

{\noindent \it  - Raw Frame Embedding Similarity (RFES).} We want to know if direct application of frame visual similarity can help reference resolution. In this baseline, the visual model $P(V|G)$ is reformulated as:
\begin{equation}
P(V|G) \propto \prod_{(i,j) \in \mathcal{A}}\prod_{\bar{z}_t = i, \bar{z}_\tau = j} s(x_t, x_\tau),
\end{equation}
where $s(\cdot,\cdot)$ is the cosine similarity between the frame embeddings given by~\cite{karpathy2015deep} and $\mathcal{A}$ is the set of all the action pairs that are connected by references in $G$. In other words, RFES model evaluates the likelihood of a graph by the total visual similarities of frames connected by the references.


{\noindent \it  - Frame Embedding Similarity (FES).} We extend RFES to FES by optimizing $s(\cdot,\cdot)$ during the M-step to maximize the probability of the current graphs.
In this case, FES is trained to help reference resolution based on frame-to-frame similarity. We compare to this baseline to understand if our model really captures fine-grained details of the image beyond frame to frame visual similarity.


{\noindent \it  - Visual+Linguistic w/o Alignment.} Our unsupervised approach faces the challenge of misaligned transcriptions and videos. We evaluate the effect of our update of $Z$ to the reference resolution task.

\begin{table}[tb]
\begin{center}
\small
\begin{tabular}[t]{ |c|c|c|c|}
  \hline			
  Methods & P & R & F1\\
  \hline
  \hline 
  Sequential Initialization & 0.483 & 0.478 & 0.480 \\
  \hline
  Random Perturbation & 0.399 & 0.386 & 0.397 \\
  \hline
  \hline
  Our Visual Model Only  & 0.294 & 0.292 & 0.293 \\\hline
  Our Linguistic Model Only \cite{kiddon2015mise} & 0.621 & 0.615 & 0.618 \\ \hline
  \hline
  RFES + Linguistic w/o Align &0.424  &0.422 & 0.423\\\hline
  FES + Linguistic w/o Align &0.547  &0.543 &0.545 \\ \hline
  Our Visual + Linguistic w/o Align &{ 0.691}	&{ 0.686} 	& { 0.688}\\ \hline
  Our Visual + Linguistic (Our Full) &{\bf 0.710}	&{\bf 0.704} &{\bf 0.707} \\
  \hline  
\end{tabular}
\end{center}
\caption{Reference resolution results. Our final model significantly outperforms the linguistic only model. Note that using vision to help reference resolution is non-trivial. Directly adding frame similarity based visual models is not improving the performance. }
\label{tab:coref}
\end{table}

\begin{figure}[tb]
\centering
   \includegraphics[width=0.99\linewidth]{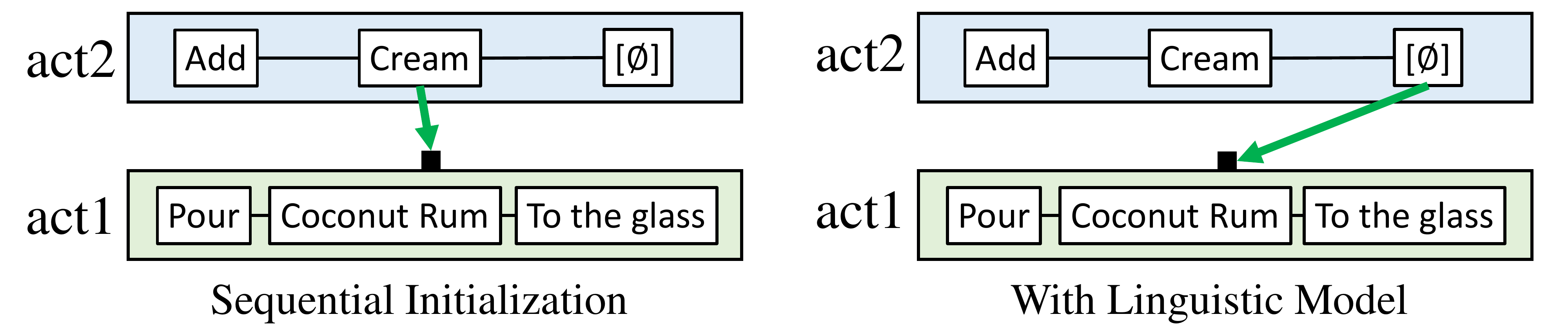} 
   \caption{Qualitative results of the linguistic model. $[\phi]$ stands for the implicit entity. On the left, the sequential baseline reference ``cream'' as the previous action outcome without understanding that it is a raw ingredient. On the other hand, our linguistic model understands (1) cream is raw ingredient, and further (2) ``add'' is usually used to combine food entities, and thus is able to infer the reference of the implicit entity correctly.
}
\label{fig:coref_qual_ling}
\end{figure}

\begin{figure}[tb]
\centering
   \includegraphics[width=0.99\linewidth]{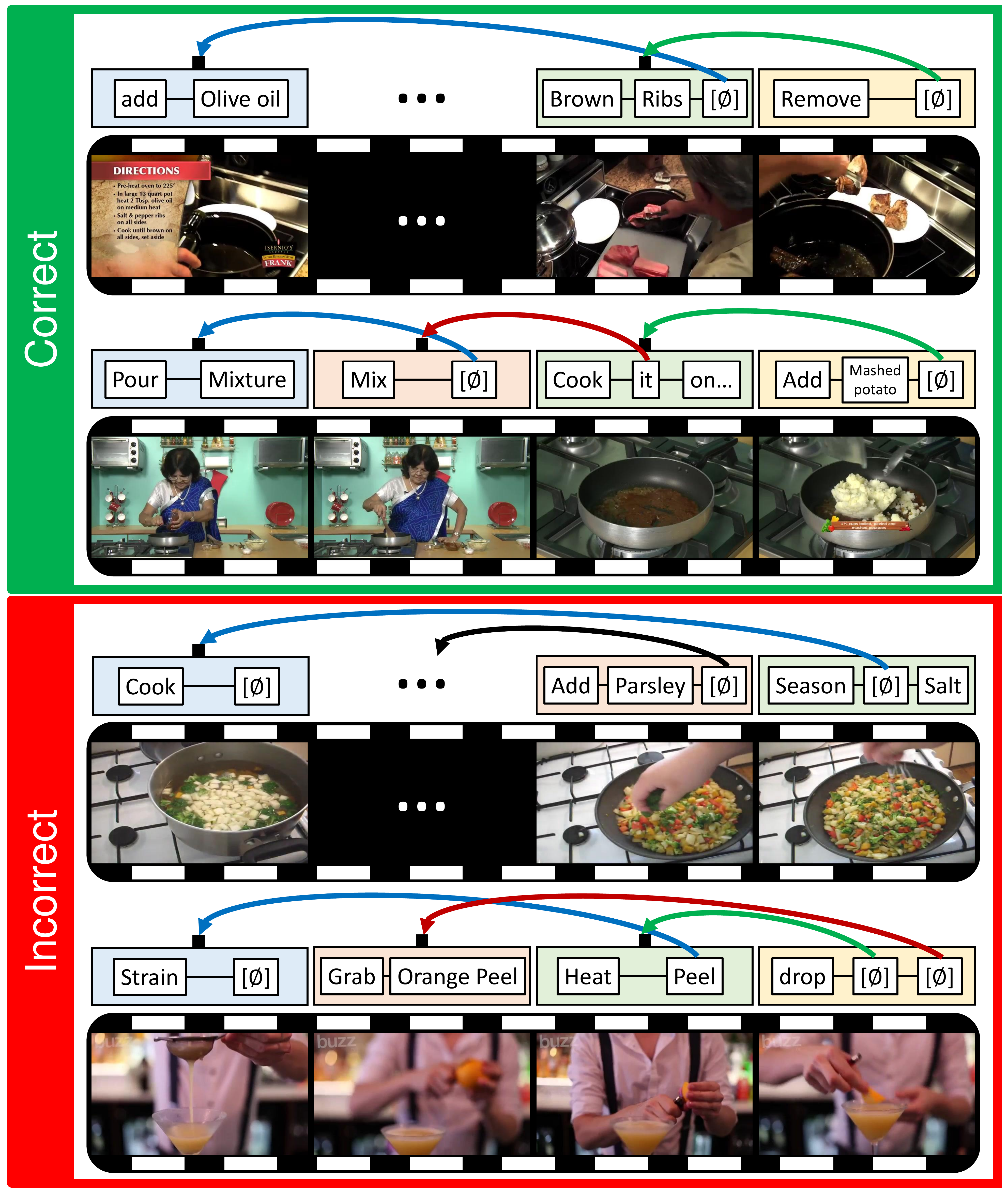} 
   \caption{Our reference resolution results. Top two rows show correct references across visual appearances. Bottom two rows show the failure cases of our model. Our visual model can sometimes be confused by similar visual appearances.}
\label{fig:coref_qual_fail}
\end{figure}

\vspace{1mm}
\noindent\textbf{Results.} 
The results are shown in \tabref{coref}. By sequential initialization, we already have a reasonable performance because of the sequential nature of instruction. This is verified by the fact that if we perform random perturbation to this graph (maximum 10 edge swaps in this case), the reference resolution performance actually goes down significantly. 
Optimizing using just the visual model for this problem, however, is not effective. Without proper regularization provided by the transcription, the visual model is unable to be trained to make reasonable reference resolution.
On the other hand, by using only our linguistic model, the performance improves over sequential baseline by resolving references including common pronoun such as ``it'', or figuring out some of the words like ``flour'' is more likely to be raw ingredients and is not referring back to previous action outcomes. Qualitative comparison of the linguistic model is shown in \figref{coref_qual_ling}. 

\noindent\textbf{Importance of our action graph embedding.}
Direct application of initial frame-level model RFES to the linguistic model, however, cannot improve the reference resolution. This is due to the visual appearance changes caused by the state changes of the entities. The extension of FES improves the performance by 10\% compared to RFES since FES optimizes the frame similarity function to help reference resolution. Nevertheless, it is still unable to improve the performance of the linguistic model because whole-frame similarity based model cannot capture fine-grained details of the graph and differentiate references from the same step. 
On the other hand, our visual model addresses both the challenge introduced by state changes and the fine-grained details of the graph by matching frames to our proposed action graph embedding. In this case, our joint visual-linguistic model further enhances the performance of linguistic model by associating the same entity across varied linguistic expressions and visual appearances that are hard to associate based on only language or frame similarity.

\noindent\textbf{Alignment can help reference resolution.}
We further verify that the joint optimization with temporal alignment $Z$ can improve the performance of our joint visual-linguistic reference resolution. In this case, as the corresponding frames are more accurate, the supervision to the visual model is less noisy and results in improved performance. Qualitative results are shown in \figref{coref_qual} to verify the improvement of both our joint visual-linguistic modeling and video-transcription alignment. \figref{coref_qual_fail} shows more qualitative results and failure cases.


\subsection{Improving Alignment by Referencing}

As discussed earlier, the alignment between captions and frames are not perfect in instructional videos. We have shown that having the alignment in the visual-linguistic model is able to improve reference resolution. In this section, we show that resolving reference is actually also beneficial to improving alignment.

Intuitively, for a sentence like ``Cut it.'', without figuring out the meaning of ``it'', it is unlikely to train a good visual model because ``it'' can be referring to food ingredient with a variety of visual appearances. This unique challenge makes the task of aligning transcription with videos more challenging compared to aligning structured text.

\begin{figure}[tb]
\centering
   \includegraphics[width=0.49\linewidth]{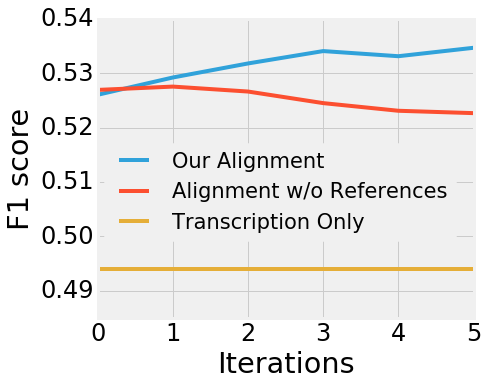} 
   \includegraphics[width=0.49\linewidth]{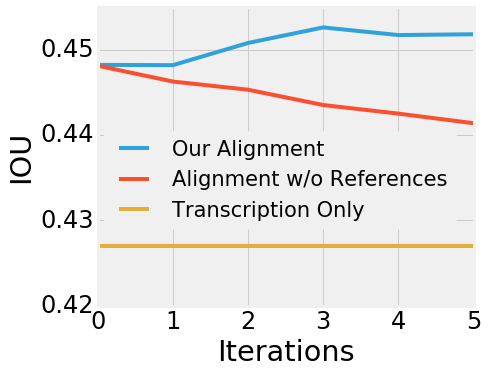}
   \caption{Video to transcription alignment results. By resolving the reference of words in the transcription, our visual-linguistic model is able to improve alignment performance over standard sentence embedding based approach. }
\label{fig:align}
\end{figure}

To verify our claim, we remove the reference resolution component in our model as the baseline. In this case, the graph embedding is reduced to the standard visual-semantic embedding of~\cite{kiros2014unifying} without the connections to the previous actions introduced by referencing. For the metrics, we follow previous works~\cite{alayrac16unsupervised,sener2015unsupervised} and use F1 score and IOU.

The alignment results are shown in \figref{align}. It can be seen that without reference resolution in the process of aligning transcription and video, the visual-semantic embedding~\cite{kiros2014unifying} is not able to improve over iterations. However, our action graph embedding resolves the references in the unstructured instructional text and is thus able to improve the alignment performance. The alignment performance of the transcription is also shown for reference.

%% file: conclusion.tex
\section{Conclusion}
We propose a new unsupervised learning approach to resolve references between actions and entities in instructional videos. Our model uses a graph representation to jointly utilize linguistic and visual models in order to handle various inherent ambiguities in videos.
Our experiments verified that our model can substantially improve upon having only one set of cues to extract meaningful references. 

%% file: egpaper_for_review.bbl
\begin{thebibliography}{10}\itemsep=-1pt

\bibitem{agrawal2016sort}
H.~Agrawal, A.~Chandrasekaran, D.~Batra, D.~Parikh, and M.~Bansal.
\newblock Sort story: Sorting jumbled images and captions into stories.
\newblock In {\em EMNLP}, 2016.

\bibitem{alayrac16unsupervised}
J.-B. Alayrac, P.~Bojanowski, N.~Agrawal, I.~Laptev, J.~Sivic, and
  S.~Lacoste-Julien.
\newblock Unsupervised learning from narrated instruction videos.
\newblock In {\em CVPR}, 2016.

\bibitem{alayrac16objectstates}
J.-B. Alayrac, J.~Sivic, I.~Laptev, and S.~Lacoste-Julien.
\newblock Joint discovery of object states and manipulating actions.
\newblock {\em arXiv:1702.02738}, 2017.

\bibitem{andreas2015alignment}
J.~Andreas and D.~Klein.
\newblock Alignment-based compositional semantics for instruction following.
\newblock In {\em EMNLP}, 2015.

\bibitem{berg2004names}
T.~L. Berg, A.~C. Berg, J.~Edwards, M.~Maire, R.~White, Y.-W. Teh,
  E.~Learned-Miller, and D.~A. Forsyth.
\newblock Names and faces in the news.
\newblock In {\em CVPR}, 2004.

\bibitem{bjorkelund2014learning}
A.~Bj{\"o}rkelund and J.~Kuhn.
\newblock Learning structured perceptrons for coreference resolution with
  latent antecedents and non-local features.
\newblock In {\em ACL}, 2014.

\bibitem{bojanowski13finding}
P.~Bojanowski, F.~Bach, I.~Laptev, J.~Ponce, C.~Schmid, and J.~Sivic.
\newblock Finding actors and actions in movies.
\newblock In {\em ICCV}, 2013.

\bibitem{bojanowski2015weakly}
P.~Bojanowski, R.~Lajugie, E.~Grave, F.~Bach, I.~Laptev, J.~Ponce, and
  C.~Schmid.
\newblock Weakly-supervised alignment of video with text.
\newblock In {\em ICCV}, 2015.

\bibitem{das2013thousand}
P.~Das, C.~Xu, R.~F. Doell, and J.~J. Corso.
\newblock A thousand frames in just a few words: Lingual description of videos
  through latent topics and sparse object stitching.
\newblock In {\em CVPR}, 2013.

\bibitem{deng2016structure}
Z.~Deng, A.~Vahdat, H.~Hu, and G.~Mori.
\newblock Structure inference machines: Recurrent neural networks for analyzing
  relations in group activity recognition.
\newblock In {\em CVPR}, 2016.

\bibitem{duchenne2009automatic}
O.~Duchenne, I.~Laptev, J.~Sivic, F.~Bach, and J.~Ponce.
\newblock Automatic annotation of human actions in video.
\newblock In {\em ICCV}, 2009.

\bibitem{durrett2013easy}
G.~Durrett and D.~Klein.
\newblock Easy victories and uphill battles in coreference resolution.
\newblock In {\em EMNLP}, 2013.

\bibitem{fidler2013sentence}
S.~Fidler, A.~Sharma, and R.~Urtasun.
\newblock A sentence is worth a thousand pixels.
\newblock In {\em CVPR}, 2013.

\bibitem{gupta2009understanding}
A.~Gupta, P.~Srinivasan, J.~Shi, and L.~S. Davis.
\newblock Understanding videos, constructing plots learning a visually grounded
  storyline model from annotated videos.
\newblock In {\em CVPR}, 2009.

\bibitem{hodosh2010cross}
M.~Hodosh, P.~Young, C.~Rashtchian, and J.~Hockenmaier.
\newblock Cross-caption coreference resolution for automatic image
  understanding.
\newblock In {\em Proceedings of the Fourteenth Conference on Computational
  Natural Language Learning}, 2010.

\bibitem{hu2016modeling}
R.~Hu, M.~Rohrbach, J.~Andreas, T.~Darrell, and K.~Saenko.
\newblock Modeling relationships in referential expressions with compositional
  modular networks.
\newblock {\em arXiv preprint arXiv:1611.09978}, 2016.

\bibitem{huang2016visual}
T.-H.~K. Huang, F.~Ferraro, N.~Mostafazadeh, I.~Misra, A.~Agrawal, J.~Devlin,
  R.~Girshick, X.~He, P.~Kohli, D.~Batra, et~al.
\newblock Visual storytelling.
\newblock In {\em NAACL}, 2016.

\bibitem{jermsurawong2015predicting}
J.~Jermsurawong and N.~Habash.
\newblock Predicting the structure of cooking recipes.
\newblock In {\em EMNLP}, 2015.

\bibitem{densecap}
J.~Johnson, A.~Karpathy, and L.~Fei-Fei.
\newblock Densecap: Fully convolutional localization networks for dense
  captioning.
\newblock In {\em Proceedings of the IEEE Conference on Computer Vision and
  Pattern Recognition}, 2016.

\bibitem{johnson2015image}
J.~Johnson, R.~Krishna, M.~Stark, L.-J. Li, D.~A. Shamma, M.~S. Bernstein, and
  L.~Fei-Fei.
\newblock Image retrieval using scene graphs.
\newblock In {\em CVPR}. IEEE, 2015.

\bibitem{karpathy2015deep}
A.~Karpathy and L.~Fei-Fei.
\newblock Deep visual-semantic alignments for generating image descriptions.
\newblock In {\em CVPR}, 2015.

\bibitem{kazemzadeh2014referitgame}
S.~Kazemzadeh, V.~Ordonez, M.~Matten, and T.~L. Berg.
\newblock Referitgame: Referring to objects in photographs of natural scenes.
\newblock In {\em EMNLP}, 2014.

\bibitem{kiddon2015mise}
C.~Kiddon, G.~T. Ponnuraj, L.~Zettlemoyer, and Y.~Choi.
\newblock Mise en place: Unsupervised interpretation of instructional recipes.
\newblock In {\em EMNLP}, 2015.

\bibitem{kiros2014unifying}
R.~Kiros, R.~Salakhutdinov, and R.~S. Zemel.
\newblock Unifying visual-semantic embeddings with multimodal neural language
  models.
\newblock {\em TACL}, 2015.

\bibitem{kiros2015skip}
R.~Kiros, Y.~Zhu, R.~R. Salakhutdinov, R.~Zemel, R.~Urtasun, A.~Torralba, and
  S.~Fidler.
\newblock Skip-thought vectors.
\newblock In {\em Advances in neural information processing systems}, pages
  3294--3302, 2015.

\bibitem{kollar2010toward}
T.~Kollar, S.~Tellex, D.~Roy, and N.~Roy.
\newblock Toward understanding natural language directions.
\newblock In {\em ACM/IEEE International Conference on Human-Robot Interaction
  (HRI)}, 2010.

\bibitem{kong2014you}
C.~Kong, D.~Lin, M.~Bansal, R.~Urtasun, and S.~Fidler.
\newblock What are you talking about? text-to-image coreference.
\newblock In {\em CVPR}, 2014.

\bibitem{krishnamurthy2013jointly}
J.~Krishnamurthy and T.~Kollar.
\newblock Jointly learning to parse and perceive: Connecting natural language
  to the physical world.
\newblock {\em TACL}, 1:193--206, 2013.

\bibitem{lau2009interpreting}
T.~A. Lau, C.~Drews, and J.~Nichols.
\newblock Interpreting written how-to instructions.
\newblock In {\em IJCAI}, 2009.

\bibitem{lee2011stanford}
H.~Lee, Y.~Peirsman, A.~Chang, N.~Chambers, M.~Surdeanu, and D.~Jurafsky.
\newblock Stanford's multi-pass sieve coreference resolution system at the
  conll-2011 shared task.
\newblock In {\em Proceedings of the Fifteenth Conference on Computational
  Natural Language Learning: Shared Task}, 2011.

\bibitem{LinCVPR14}
D.~Lin, S.~Fidler, C.~Kong, and R.~Urtasun.
\newblock Visual semantic search: Retrieving videos via complex textual
  queries.
\newblock In {\em CVPR}, 2014.

\bibitem{liu2016jointly}
C.~Liu, S.~Yang, S.~Saba-Sadiya, N.~Shukla, Y.~He, S.-C. Zhu, and J.~Y. Chai.
\newblock Jointly learning grounded task structures from language instruction
  and visual demonstration.
\newblock In {\em EMNLP}, 2016.

\bibitem{long2016simpler}
R.~Long, P.~Pasupat, and P.~Liang.
\newblock Simpler context-dependent logical forms via model projections.
\newblock In {\em ACL}, 2016.

\bibitem{maeta2015framework}
H.~Maeta, T.~Sasada, and S.~Mori.
\newblock A framework for procedural text understanding.
\newblock In {\em Proceedings of the 14th International Conference on Parsing
  Technologies}, 2015.

\bibitem{malmaud2015s}
J.~Malmaud, J.~Huang, V.~Rathod, N.~Johnston, A.~Rabinovich, and K.~Murphy.
\newblock What’s cookin’? interpreting cooking videos using text, speech
  and vision.
\newblock In {\em NAACL HLT}, 2015.

\bibitem{malmaud2014cooking}
J.~Malmaud, E.~J. Wagner, N.~Chang, and K.~Murphy.
\newblock Cooking with semantics.
\newblock In {\em Proceedings of the ACL 2014 Workshop on Semantic Parsing},
  2014.

\bibitem{manning2014stanford}
C.~D. Manning, M.~Surdeanu, J.~Bauer, J.~R. Finkel, S.~Bethard, and
  D.~McClosky.
\newblock The stanford corenlp natural language processing toolkit.
\newblock In {\em ACL (System Demonstrations)}, pages 55--60, 2014.

\bibitem{mao2015generation}
J.~Mao, J.~Huang, A.~Toshev, O.~Camburu, A.~Yuille, and K.~Murphy.
\newblock Generation and comprehension of unambiguous object descriptions.
\newblock In {\em CVPR}, 2016.

\bibitem{martschat2015latent}
S.~Martschat and M.~Strube.
\newblock Latent structures for coreference resolution.
\newblock {\em TACL}, 3:405--418, 2015.

\bibitem{mikolov2013distributed}
T.~Mikolov, I.~Sutskever, K.~Chen, G.~S. Corrado, and J.~Dean.
\newblock Distributed representations of words and phrases and their
  compositionality.
\newblock In {\em NIPS}, 2013.

\bibitem{nagaraja2016modeling}
V.~K. Nagaraja, V.~I. Morariu, and L.~S. Davis.
\newblock Modeling context between objects for referring expression
  understanding.
\newblock In {\em ECCV}, 2016.

\bibitem{naim2015discriminative}
I.~Naim, Y.~C. Song, Q.~Liu, L.~Huang, H.~Kautz, J.~Luo, and D.~Gildea.
\newblock Discriminative unsupervised alignment of natural language
  instructions with corresponding video segments.
\newblock {\em NAACL HLT}, 2015.

\bibitem{pennington2014glove}
J.~Pennington, R.~Socher, and C.~D. Manning.
\newblock Glove: Global vectors for word representation.
\newblock In {\em EMNLP}, 2014.

\bibitem{pirsiavash2014parsing}
H.~Pirsiavash and D.~Ramanan.
\newblock Parsing videos of actions with segmental grammars.
\newblock In {\em CVPR}, 2014.

\bibitem{plummer2015flickr30k}
B.~A. Plummer, L.~Wang, C.~M. Cervantes, J.~C. Caicedo, J.~Hockenmaier, and
  S.~Lazebnik.
\newblock Flickr30k entities: Collecting region-to-phrase correspondences for
  richer image-to-sentence models.
\newblock In {\em ICCV}, 2015.

\bibitem{qi2009efficient}
G.-J. Qi, J.~Tang, Z.-J. Zha, T.-S. Chua, and H.-J. Zhang.
\newblock An efficient sparse metric learning in high-dimensional space via l
  1-penalized log-determinant regularization.
\newblock In {\em ICML}, 2009.

\bibitem{ramanathan2014linking}
V.~Ramanathan, A.~Joulin, P.~Liang, and L.~Fei-Fei.
\newblock Linking people in videos with “their” names using coreference
  resolution.
\newblock In {\em ECCV}, 2014.

\bibitem{ramanathan2013video}
V.~Ramanathan, P.~Liang, and L.~Fei-Fei.
\newblock Video event understanding using natural language descriptions.
\newblock In {\em ICCV}, 2013.

\bibitem{rohrbach2015grounding}
A.~Rohrbach, M.~Rohrbach, R.~Hu, T.~Darrell, and B.~Schiele.
\newblock Grounding of textual phrases in images by reconstruction.
\newblock In {\em ECCV}, 2016.

\bibitem{schlangen2009incremental}
D.~Schlangen, T.~Baumann, and M.~Atterer.
\newblock Incremental reference resolution: The task, metrics for evaluation,
  and a bayesian filtering model that is sensitive to disfluencies.
\newblock In {\em SIGDIAL}, 2009.

\bibitem{schuster2015generating}
S.~Schuster, R.~Krishna, A.~Chang, L.~Fei-Fei, and C.~D. Manning.
\newblock Generating semantically precise scene graphs from textual
  descriptions for improved image retrieval.
\newblock In {\em Proceedings of the Fourth Workshop on Vision and Language},
  2015.

\bibitem{sener2015unsupervised}
O.~Sener, A.~R. Zamir, S.~Savarese, and A.~Saxena.
\newblock Unsupervised semantic parsing of video collections.
\newblock In {\em ICCV}, 2015.

\bibitem{sigurdsson2016learning}
G.~A. Sigurdsson, X.~Chen, and A.~Gupta.
\newblock Learning visual storylines with skipping recurrent neural networks.
\newblock In {\em ECCV}, 2016.

\bibitem{socher2014grounded}
R.~Socher, A.~Karpathy, Q.~V. Le, C.~D. Manning, and A.~Y. Ng.
\newblock Grounded compositional semantics for finding and describing images
  with sentences.
\newblock {\em TACL}, 2:207--218, 2014.

\bibitem{soran2015generating}
B.~Soran, A.~Farhadi, and L.~Shapiro.
\newblock Generating notifications for missing actions: Don't forget to turn
  the lights off!
\newblock In {\em ICCV}, 2015.

\bibitem{tellex2014learning}
S.~Tellex, P.~Thaker, J.~Joseph, and N.~Roy.
\newblock Learning perceptually grounded word meanings from unaligned parallel
  data.
\newblock {\em Machine Learning}, 94(2):151--167, 2014.

\bibitem{vinyals2015show}
O.~Vinyals, A.~Toshev, S.~Bengio, and D.~Erhan.
\newblock Show and tell: A neural image caption generator.
\newblock In {\em Proceedings of the IEEE Conference on Computer Vision and
  Pattern Recognition}, 2015.

\bibitem{xiong2016robot}
C.~Xiong, N.~Shukla, W.~Xiong, and S.-C. Zhu.
\newblock Robot learning with a spatial, temporal, and causal and-or graph.
\newblock In {\em ICRA}, 2016.

\bibitem{xu2015jointly}
R.~Xu, C.~Xiong, W.~Chen, and J.~J. Corso.
\newblock Jointly modeling deep video and compositional text to bridge vision
  and language in a unified framework.
\newblock In {\em AAAI}, 2015.

\bibitem{yang2016grounded}
S.~Yang, Q.~Gao, C.~Liu, C.~Xiong, S.-C. Zhu, and J.~Y. Chai.
\newblock Grounded semantic role labeling.
\newblock In {\em Proceedings of NAACL-HLT}, 2016.

\bibitem{yu2016modeling}
L.~Yu, P.~Poirson, S.~Yang, A.~C. Berg, and T.~L. Berg.
\newblock Modeling context in referring expressions.
\newblock In {\em ECCV}, 2016.

\bibitem{yu2014instructional}
S.-I. Yu, L.~Jiang, and A.~Hauptmann.
\newblock Instructional videos for unsupervised harvesting and learning of
  action examples.
\newblock In {\em ACM MM}, 2014.

\bibitem{zhu2015aligning}
Y.~Zhu, R.~Kiros, R.~Zemel, R.~Salakhutdinov, R.~Urtasun, A.~Torralba, and
  S.~Fidler.
\newblock Aligning books and movies: Towards story-like visual explanations by
  watching movies and reading books.
\newblock In {\em ICCV}, 2015.

\bibitem{zitnick2013bringing}
C.~L. Zitnick and D.~Parikh.
\newblock Bringing semantics into focus using visual abstraction.
\newblock In {\em CVPR}, 2013.

\end{thebibliography}
